\documentclass{article}

\usepackage{microtype}
\usepackage{graphicx}
\usepackage{subfigure}
\usepackage{booktabs} 
\usepackage{amsfonts}       
\usepackage{nicefrac}       
\usepackage{microtype}      
\usepackage{threeparttable}
\usepackage{algorithm}
\usepackage{algorithmic}
\usepackage{subfigure}
\usepackage{amsmath, amsfonts, amssymb}
\usepackage{hyperref}
\usepackage{graphicx}
\usepackage{tikz}
\usetikzlibrary{shapes,snakes}

\usepackage{multirow}
\usepackage{fixmath}
\usepackage{paralist}
\usepackage{footnote}
\usepackage{wrapfig}
\usepackage{booktabs,dcolumn}

\newtheorem{theorem}{Theorem}

\newtheorem{lemma}[theorem]{Lemma}
\newtheorem{corollary}{Corollary}

\newtheorem{proposition}{Proposition}

\renewcommand{\algorithmicrequire}{\textbf{Input:}}  
\renewcommand{\algorithmicensure}{\textbf{Output:}} 
\DeclareMathOperator*{\argmin}{arg\,min}

\usepackage{hyperref}

\usepackage[accepted]{icml2018}

\icmltitlerunning{Fast Variance Reduction Method with Stochastic Batch Size}

\begin{document}

\twocolumn[
\icmltitle{Fast Variance Reduction Method with Stochastic Batch Size}
\icmlsetsymbol{equal}{*}

\begin{icmlauthorlist}
\icmlauthor{Xuanqing Liu}{cs}
\icmlauthor{Cho-Jui Hsieh}{cs,stat}
\end{icmlauthorlist}

\icmlaffiliation{cs}{Department of Computer Science, University of California, Davis, California, USA}
\icmlaffiliation{stat}{Department of Statistic, University of California, Davis, California, USA}

\icmlcorrespondingauthor{Xuanqing Liu}{xqliu@ucdavis.edu}
\icmlcorrespondingauthor{Cho-Jui Hsieh}{chohsieh@ucdavis.edu}

\icmlkeywords{Machine Learning, ICML}

\vskip 0.3in
]

\printAffiliationsAndNotice{} 

\begin{abstract}
In this paper we study a family of variance reduction methods with randomized batch size---at each step, the algorithm first randomly chooses the batch size and then selects a batch of samples to conduct a variance-reduced stochastic update. We give the linear convergence rate for this framework for composite functions, and show that the optimal strategy to achieve the optimal convergence rate per data access is to always choose batch size of 1, which is equivalent to the SAGA algorithm.
However, due to the presence of cache/disk IO effect in computer architecture, the number of data access cannot reflect the running time because of 1) random memory access is much slower than sequential access, 2) when data is too big to fit into memory, disk seeking takes even longer time. After taking these into account, choosing batch size of $1$ is no longer optimal, so we propose a new algorithm called SAGA++ and show how to calculate the optimal average batch size
    theoretically. Our algorithm outperforms
SAGA and other existing batched and stochastic solvers on real datasets. 
In addition, we also conduct a precise analysis to compare different update rules for variance reduction methods, 
showing that SAGA++ converges faster than SVRG in theory.
\end{abstract}
\section{Introduction}
In this paper, we consider the following finite-sum composite optimization problem: 
\begin{equation}\label{problem}
w^*=\argmin_{w\in\mathbb{R}^p} \Big\{ F(w)\triangleq\frac{1}{n}\sum_{i=1}^n f_i(w)+g(w) \Big\}. 
\end{equation}
Here we assume each $f_i(w)$ is a $\mu$-strongly convex, $L$-smooth function, the regularization term $g(w)$ is convex but not necessarily differentiable. In machine learning applications, $n$ is the number of training samples, 
each $f_i$ is the loss function such as logistic loss or $\ell_2$ loss, and $g(w)$ is the regularization term which can be non-smooth (e.g., $\ell_1$ regularization). 
For large data, SGD is preferred over gradient descent and has been widely used in large-scale applications. However, since the variance of stochastic gradient will not go to zero even when $w=w^*$ (the optimal solution), SGD has to gradually shrink the step size to guarantee convergence, at the cost of suboptimal rate. 
To speed up the convergence, there is a recent line of research on developing new algorithms with linear convergence rate using variance reduction techniques, the representative work includes SAG~\cite{hofmann2015variance}, SVRG~\cite{johnson2013accelerating}, SAGA~\cite{defazio2014saga}, S2GD~\cite{konecny2013semi} etc. Further, one can accelerate this framework via concepts similar to the Nesterov's momentum
method~\cite{lin2015universal,allen2017katyusha}. 

The motivation of this paper is to study the effect of batch size in variance reduction methods. 
The effect of batch size in SGD (without variance reduction) has been studied in the literature such as \cite{li2014efficient, bengio2012practical, keskar2016large}. 
Assuming a subset of $b$ samples is chosen for SGD at each step, the theoretical analysis in~\cite{dekel2012optimal} suggests that the error
is at the order of $\mathcal{O}(1/\sqrt{bT}+1/T)$ after $T$ iterations, 
and this bound is later improved to $\mathcal{O}(1/\sqrt{bT})$ in \cite{li2014efficient}. When constraining on SVM-hinge loss, \cite{takac2013mini} also shows an order of $\mathcal{O}(\frac{n}{b}+\frac{\beta_b}{b}\cdot\frac{1}{\lambda\epsilon})$ iterations to get an $\epsilon$-suboptimal solution. Since
each iteration will take the time proportional to $b$, these bounds suggest that the acceleration
of convergence exactly covers the overhead of each iteration. It is thus interesting to see whether the same
conclusion also applies for variance reduction methods.


To answer this question, we study a family of variance reduction methods with randomized batch sizes. 
At each iteration, the algorithm  first  randomly selects the batch size and then chooses a batch of samples to conduct
a variance reduced stochastic update. Our main findings and contributions can be listed as follows:  
\begin{compactitem}
    \item We prove linear convergence rate for this family of stochastic batched variance reduction algorithms. Our result covers composite minimization problems with non-smooth regularizations, and any distribution of batch sizes. 
    \item Interestingly, with this unified analysis, we theoretically show that the convergence rate can be maximized if the algorithm always chooses batch size of 1. Therefore, increasing batch size does not help in terms of the number of data access. 
    \item However, the number of data access does not precisely reflect the actual running time due to the memory hierarchy and cache/disk IO effect in computer architectures---accessing a continuous block of memory is faster than accessing disjoint ones, and disk seeking costs even more. After taking these into account, we propose the SAGA++ algorithm, and show how to calculate the optimal average batch size in practice. Our algorithm outperforms existing algorithms in terms of running time. 
    \item In addition, we also develop a more precise analysis for comparing the convergence rates of variance reduction methods, and develop an algorithm to universally accelerate the stochastic methods for solving $\ell_1$-regularized problems by lazy updates. Which rediscovers \cite{konevcny2016mini} independently.
    \end{compactitem}

\paragraph{Related Work}
We will discuss the related variance reduction methods in next section. Here we 
describe some other related work on stochastic optimization. 

Stochastic optimization has become popular due to their vast and far reaching applications in large-scale machine learning, and this is also one of our main focus in this paper. Among them, stochastic gradient descent has been widely used, and its variants~\cite{duchi2011adaptive,kingma2014adam}
are popular for training deep neural networks. 
There are also other examples, such as stochastic coordinate descent~\cite{nesterov2012efficiency} and stochastic dual coordinate descent~\cite{shalev2013stochastic}. 
At each iteration, SGD selects one sample to conduct the update, but its gradient often contains huge noise. To reduce the noise or variance, mini-batch SGD has been intensively studied in the literature, including~\cite{li2014efficient}, and the recent work on big-batch SGD~\cite{de2016big}. Some theoretical results have been discussed in our introduction~\cite{li2014efficient, dekel2012optimal}.

Although some recent works have discussed about mini-batch variance reduction algorithms~\cite{hofmann2015variance,harikandeh2015stopwasting}, there is no clear conclusion on whether increasing the batch size helps the convergen speed. Ideally the convergence rate $1-\rho$ should linearly depend on the batch size: $\rho\propto b$; if that is the case, simply by calculating the batched gradient in parallel we will see linear speed up. \cite{hofmann2015variance} suggests $\rho\approx b/n$ in big
data regime and $\rho$ is independent of $b$ in ill-conditioned case, this can be regarded as an asymptotic situation of our result, which claims that $\rho$ is a increasing function of $b$, but a larger batch size is less useful when the Hessian is ill conditioned. However, with a more precise bound in terms of $b$, we are able to show that $b=1$ is always optimal in terms of number of data access. 
\par
As to the sampling techniques, the random sampling of batch size is seen in \cite{richtarik2016parallel} where the authors considered about the partially separable functions and apply block coordinate descent by randomly generate a set of blocks with arbitrary size. Similar idea is later exploited in~\cite{qu2015quartz,csiba2015primal}. Our idea differs from these previous works in that we put computer architecture effects into account when deciding whether we should choose full gradient or stochastic gradient to update parameters.

\section{Framework: Variance Reduction with Stochastic Batch Size}
\label{sec:framework}

Our proposed framework is shown in Algorithm \ref{alg:batch_method}: at each iteration, the algorithm first randomly chooses the batch size, ranging from $1$ to $n$, and then samples a batch accordingly. 
We use $\mathcal{B}$ as a random set to denote the mini-batch chosen at each step, and its batch size, denoted as $|\mathcal{B}|$, is a random variable. Denote $f'_i(\phi_i^t)$ as the previous gradient evaluated on sample $x_i$ and $w^t$ 
is the iterate at time $t$. 
The update rule is given by:
\begin{equation}
\label{update}
w^{t+1}=\mathsf{Prox}_{\gamma g(\cdot)}\big(w^t-\gamma G(w^t)\big), 
\end{equation}
where $\gamma$ is the step size and $G(w^t)$ is the unbiased gradient estimator:
\begin{equation}
\label{gradient}
G(w^t)=\frac{1}{|\mathcal{B}|}\sum_{i\in\mathcal{B}}f_i'(w^t)\underbrace{-\frac{1}{|\mathcal{B}|}\sum_{i\in\mathcal{B}}f_i'(\phi_i^t)
+ \bar{u}}_{\text{control variate}}, 
\end{equation}
where $\bar{u} = \frac{1}{n}\sum_{i=1}^n f_i'(\phi_i^t)$ is stored and maintained in memory. 
Similar to SAGA, in general the algorithm needs to store all the vectors $f_i'(\phi_i^t)$ in memory, 
but for many commonly used cases it only needs to store a scalar for each sample index $i$. 
For example, in GLM problems where $f_i(w)=\ell_i(x_i^\intercal w)$, 
since $f_i'(w) = x_i \ell_i'(x_i^\intercal w)$ we only need to store a scalar $\ell_i'(x_i^\intercal \phi_i^t)$ for
each $i\in\{1,2,\dots, n\}$. 
\begin{algorithm}[tb]
\caption{Variance Reduction Method with Stochastic Batch Size}
\label{alg:batch_method}
\begin{algorithmic}
\STATE{\bfseries Input:} training samples $\{(x_i, y_i)\}_{i=1}^n$, initial guess $w_0$
\STATE{\bfseries Output:} $w^*=\arg\min_w F(w)$
\STATE $w=w_0$;
\FOR{\texttt{iter}$=0$ {\bfseries to } \texttt{MAX\_ITER}}
\STATE Choose a batch size $1\le b\le n$ randomly based on some distribution;
\STATE Sample a batch $\mathcal{B}\subseteq \{1,2,\dots,n\}$, $|\mathcal{B}|=b$;
\STATE Calculate variance reduced gradient vector by \eqref{gradient};
\STATE Apply update according to \eqref{update};
\STATE Update gradient memory: $f'_i(\phi_i^{t+1})\leftarrow f'_i(w^t)$, $\forall i\in \mathcal{B}$;
\STATE Update $\bar{u} \leftarrow \bar{u} + \frac{1}{n}\sum_{i\in \mathcal{B}}f'_i(w_t) - \frac{1}{n} \sum_{i\in\mathcal{B}}f'_i(\phi_i^{t})$
\ENDFOR
\STATE Return $w^*=w$;
\end{algorithmic}
\end{algorithm}

Algorithm \ref{alg:batch_method} is very general and can include most of the existing variance reduction methods because our algorithm does not put any restriction on the choice of batch size, which can be either random or fixed. We discuss the connections between this framework and others: 
\begin{compactitem}
    \item When the batch size is $n$ with probability $1$, 
    Algorithm~\ref{alg:batch_method} will compute the full gradient at each iteration, which is equivalent to gradient descent. 
    \item When the batch size is always $1$, the algorithm is equivalent to SAGA~\cite{defazio2014saga}. At each step, SAGA uniformly chooses one sample from $\{1,2,\dots,n\}$ and then update the iterates by the same variance reduced gradient defined in~\eqref{gradient}. 
    \item SVRG~\cite{johnson2013accelerating}: This method adopts two layers of iterations. In each outer iteration SVRG calculates full gradient (also called gradient snapshot), and in each inner iteration it chooses one sample to update. 
    SVRG does not update the gradient snapshot inside the inner iteration, so strictly speaking it cannot fit into our framework. However our algorithm, SAGA++, based partly on SVRG adopts a better update rule which will be discussed later.
    \item S2GD, mS2GD: They are variants of SVRG when the number of inner iterations $m$ follows a probability distribution: $m\sim (1-\nu \gamma)^{M-m}/\beta$, $m=1,2,\dots,M$ where $\nu$ is the lower bound of strongly convex factor $\mu$, $\beta$ is a normalizing factor and $\gamma$ is step size. 
     \cite{konevcny2016mini} extends S2GD to mini-batched version. 
\end{compactitem}

\section{Theoretical Analysis and New Algorithms}
\label{sec:theory}
We discuss our theoretical results and new insights in this section. 
First, we prove the linear convergence rate of Algorithm~\ref{alg:batch_method} in Section~\ref{sec:convergence}, and then in Section~\ref{sec:saga++} we will take the cache/disk IO effect into consideration
to derive the new algorithm SAGA++. We show the SAGA-style update
used in this paper is more efficient than SVRG-style update in Section~\ref{sec:onestep} and then 
discuss a new technique to conduct lazy update for $\ell_1$ regularization in our Algorithm~\ref{alg:batch_method}. 
We left the proof in appendix.
\subsection{Convergence rate analysis}
\label{sec:convergence}
We assume the objective function is $\mu$-strongly convex and $L$-Lipschitz smooth, 
and $\kappa=L/\mu$ is the condition number. 
We will use the following useful bounds in our analysis:
\begin{subequations}
\begin{align}
&f(y)\ge f(x)+f'(x)(y-x)+\mu/2\Vert y-x\Vert^2\\
&f(y)\le f(x)+f'(x)(y-x)+L/2\Vert y-x \Vert^2. 
\end{align}
\end{subequations}
Hereafter we use $\Vert\cdot\Vert$ to denote $\ell_2$ norm unless stated explicitly.
To simplify notation, we define $f_i^{\delta}(w)=f_i(w)-f_i(w^*)- f_i'(w^*)(w-w^*)$ and $f^{\delta}(w)=\frac{1}{n}\sum_{i=1}^nf^{\delta}_i(w)$ as the Bregman divergence between $w$ and $w^*$, and define $\bar{H}_t=\frac{1}{n}\sum_{i=1}^n f_i^{\delta}(\phi_i^t)$ 
to represent the averaged Bregman divergence between $w^*$ and the snapshots $\phi_i^t$. 

To show the convergence of Algorithm~\ref{alg:batch_method}, we first calculate the expected change of $\bar{H}_t$ 
after each update: 

\begin{lemma}
\label{le:H}
For the update rule \eqref{update} and $\bar{H}_t$ defined above, we have 
$
\mathbb{E}\bar{H}_{t+1}=\frac{n-\mathbb{E}|\mathcal{B}|}{n}\bar{H}_t+\frac{\mathbb{E}|\mathcal{B}|}{n}f^{\delta}(w^t). 
$
\end{lemma}
 Note that unlike \cite{hofmann2015variance} where $|\mathcal{B}|$ is deterministic, here we generalize their result to allow random batch size. Similarly, the progress of $\Vert w^t-w^* \Vert$ can be bounded by:
\begin{lemma}
\label{le:distance}
Define iteration progress by the distance to the optimal solution $w^*$, then:
\begin{align}\label{eq:distance}
\Vert w^{t}-w^*\Vert^2-\mathbb{E}\Vert w^{t+1}-w^*\Vert^2 \ge \gamma\mu\Vert w^t-w^* \Vert^2\\\nonumber
- ((1+\beta)\gamma^2-\gamma/L)\mathbb{E}\Vert f_i'(w^t)-f'_i(w^*) \Vert^2 \\\nonumber
+ \gamma^2\beta\Vert f'(w^t)-f'(w^*) \Vert^2+\frac{2\gamma(L-\mu)}{L}f^{\delta}(w^t)\\\nonumber -2(1+\beta^{-1})L\gamma^2 \bar{H}_t,
\end{align}
where $\beta>0$ is an arbitrary constant.
\end{lemma}
Combining Lemma~\ref{le:H} with \ref{le:distance}, we can build a contraction on Lyapunov function as follows.
\begin{theorem}
\label{th:Lyapunov}
Define Lyapunov function as $\mathcal{L}_t=c\bar{H}_t+\Vert w^t-w^* \Vert^2$,  $c$ is a predefined constant, then we have  $\mathbb{E}\mathcal{L}_{t+1}\le (1-\rho)\mathcal{L}_t$ where $\rho=\min(\frac{\mathbb{E}|\mathcal{B}|}{n}-\frac{2(1+\beta^{-1})L\gamma^2}{c}, \gamma\mu)$ if 
the step size $\gamma$ satisfies the following conditions:
\begin{equation}\label{eq:gamma_bound}
\begin{aligned}
&\text{Upper bound}:\\
&\quad\gamma\le \min\Big(\frac{1}{(1+\beta)L}, \sqrt{\frac{c\mathbb{E}|\mathcal{B}|}{2(1+\beta^{-1})nL}}\Big).\\
&\text{Lower bound}:\\
&\quad 2\mu\beta\gamma^2+\frac{2(L-\mu)}{L}\gamma-\frac{c\mathbb{E}|\mathcal{B}|}{n}\ge 0.
\end{aligned}
\end{equation}
Recall $c$, $\beta$ are predefined constants.
\end{theorem}
However, the result above is too complex to interpret. To get a better understanding of how the averaged batch size $\mathbb{E}|\mathcal{B}|$, step size $\gamma$ and contraction factor $\rho$ are related to each other, we simplify the result along different directions.
\begin{proposition}(\textbf{Adaptive step size}) In this case we want the step size $\gamma$ to be independent on strong convexity $\mu$. To this end, we set $\beta=2$, $c=\frac{n}{3L\mathbb{E}|\mathcal{B}|}$, so that $\gamma=\frac{1}{3L}$.
\end{proposition}
This result is the same with the adaptive step size of SAGA, which trade simplicity with tightness (the step size and convergence rate is independent on $\mathbb{E}|\mathcal{B}|$ so we can not see the benefit of larger batch). To develop a more informative result, we resort to the following proposition:
\begin{proposition}(\textbf{$\mathbb{E}|\mathcal{B}|$-dependent step size})
\label{pr:convergence}
If we set step size to  $\gamma=\frac{c}{8\kappa}\Big(\sqrt{1+\frac{16\kappa\mathbb{E}|\mathcal{B}|}{cn\mu}}-1\Big)$
and $c=\frac{\tau n}{L\mathbb{E}|\mathcal{B}|}$, $\tau\in (0,1)$ is a constant, our algorithm converges linearly with a contraction factor $1-\rho$, i.e. $\Vert w^{t}-w^*\Vert^2 \le (1-\rho)^t\big[\Vert w^0-w^*\Vert^2+c\bar{H}_0\big]$ and $\rho=\gamma\mu$.
\end{proposition}
The selection of step size in Proposition \ref{pr:convergence} is optimal in terms of maximizing the convergence rate, as proven in appendix. Admittedly, after developing these results, the convergence rate and step size are loose after many inequalities, so these bounds should be regarded as the worst case situation. Even so, as a quick verification, we can show that our result matches the bounds of gradient descent and SAGA in the following extreme cases: 
\begin{itemize}
    \item Gradient descent: when setting $\mathbb{P}(|\mathcal{B}|=n)=1$, then $\rho_{\text{GD}}=\frac{\tau}{8\kappa^2 }(\sqrt{16\kappa^2/\tau+1}-1)\approx \frac{\sqrt{\tau}}{2\kappa}$ this gives the same order as the standard rate of gradient descent ($\frac{2\mu}{L+\mu}$). 
    
    \item SAGA: $\mathbb{P}(|\mathcal{B}|=1)=1$, for the ill-conditioned case 
    where $\kappa$ is comparable to $n$, $\rho=\frac{1}{\kappa}\cdot\text{constant}\approx O(\frac{1}{\kappa})$. While in the well-conditioned case, $\kappa\ll n$, we have $\rho=\frac{\tau n}{8\kappa^2}(\sqrt{1+\frac{16\kappa^2}{\tau n^2}}-1)\approx O(\frac{1}{n})$. These rates match the results in the original SAGA paper~\cite{defazio2014saga}.
    
\end{itemize}
Note that the step size, unlike SGD, is always bounded away from zero (we do not need to decrease the step size in each epoch). This can be seen from the fact that $\gamma=\frac{c}{8\kappa}\Big(\sqrt{1+\frac{16\kappa\mathbb{E}|\mathcal{B}|}{cn\mu}}-1\Big)\ge 0$. 

Next we try to find out the optimal averaged batch size $\mathbb{E}|\mathcal{B}|$ in order to achieve the optimal convergence rate in terms of {\it number of data access}. Based on Proposition~\ref{pr:convergence}, to return an $\epsilon$-accurate solution we need
$\mathcal{O}(\frac{\log \epsilon}{\log(1-\rho)})$ iterations, which implies $\mathcal{O}(\frac{\log\epsilon}{\log(1-\rho)}\mathbb{E}|\mathcal{B}|)$ epochs if the averaged batch size is $\mathbb{E}|\mathcal{B}|$. We then derive the following corollary to show that simply increasing the batch size will slow down the convergence rate per data access: 
\begin{corollary}\label{Coro:theory_B}
(\textbf{Theoretically optimal batch size}) Since the effective number of data access per iteration is proportional to the averaged batch size $\mathbb{E}|\mathcal{B}|$, the optimal batch size should maximize the decrement of function value with fixed gradient calculation, which can be formulated as follows:
\begin{equation}\label{optimal_B}
\begin{aligned}
\mathbb{E}|\mathcal{B}|^*&=\argmin_{\mathbb{E}|\mathcal{B}|}\frac{\log\epsilon}{\log (1-\rho)}\mathbb{E}|\mathcal{B}|\approx\argmin_{\mathbb{E}|\mathcal{B}|}\frac{\mathbb{E}|\mathcal{B}|}{\rho}\\
&=\argmin_{\mathbb{E}|\mathcal{B}|}\frac{\mathbb{E}|\mathcal{B}|^2}{\sqrt{1+\frac{16\kappa^2}{\tau n^2}\mathbb{E}|\mathcal{B}|^2}-1}.
\end{aligned}
\end{equation}
By taking derivative it is easy to see that the function is monotone increasing with $\mathbb{E}|\mathcal{B}|$ (we leave it in appendix). So theoretically $\mathbb{E}|\mathcal{B}|=1$ (which is SAGA method) is optimal.
\end{corollary}

\subsection{SAGA++: Optimal batch sizes when taking cache/disk IO effect into consideration}
\label{sec:saga++}

\begin{figure*}
    \centering
    \includegraphics[width=0.7\linewidth]{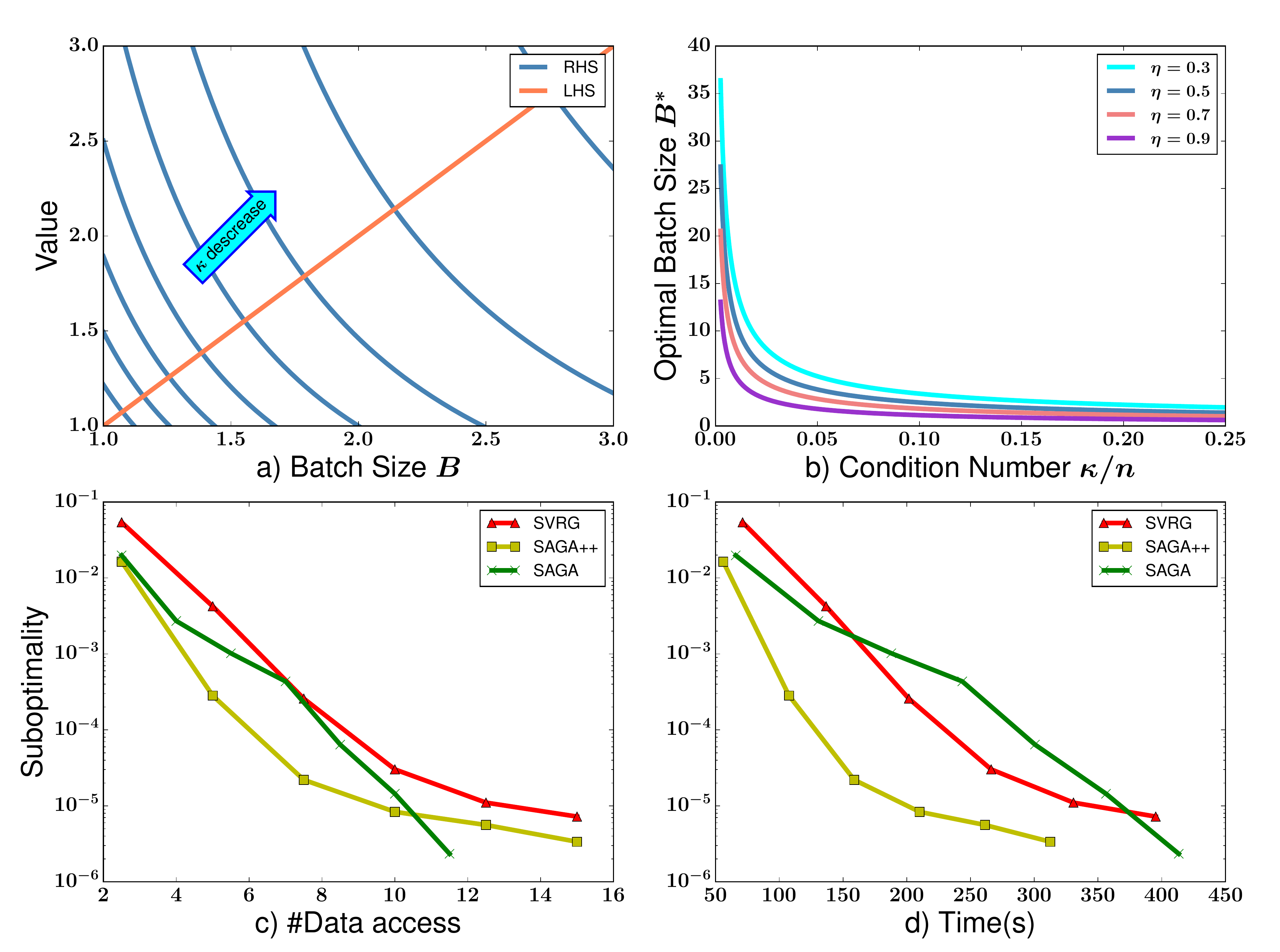}
    \caption{a) solution of \eqref{optimal_B_equation} when cache/disk IO effect coefficient $\eta=0.7$, the optimal batch size $\mathbb{E}\mathcal{B}^*$ is the intersection of two lines (marked as blue and orange), in this plot the condition number ranges from $0.025n$ to $0.5n$. b) To see the relation of $\mathbb{E}B^*$ and condition number $\kappa$ more clearly, we  solve \eqref{optimal_B_equation} numerically, the optimal batch size drop rapidly when $\kappa$ grows. At the same level of condition number, we should use a larger average batch when the cache/disk IO effect is strong ($\eta$ small). c,d) Experiment on \texttt{avazu} dataset(cache/disk IO effect ratio $\eta=0.46$), with respect to both data access(gradient calculation) and running time. }
    \label{fig:optimal_B}
\end{figure*}

According to Corollary~\ref{Coro:theory_B}, one should always choose $|\mathcal{B}|=1$ in order to minimize the number of data accesses. However, small number of data access may not necessarily lead to short running time
in practice---in modern computer architectures, ``sequential accesses'' of data stored in the memory is much faster than ``random accesses'', because accessing the memory wildly can result in frequent cache miss or disk seeking. Therefore, calculating the full gradient $f'(w)$ will take less time than calculating $n$ random gradient components $f'_i(w)$ (see Table~\ref{tab:datasets} for measurement result). This leads to a new variance reduction method with non-deterministic batch
size selection strategy(SAGA++) that combines full gradient access and SAGA ($|\mathcal{B}|=1$): 
at each step we choose $|\mathcal{B}|=n$ with probability $p$ and $|\mathcal{B}|=1$ with probability $1-p$. 
When $|\mathcal{B}|=n$, SAGA++ 
accesses the whole dataset and this can be relatively fast due to the sequential memory access pattern, 
while when $|\mathcal{B}|=1$ it randomly accesses one sample. 
By changing $p$ we can smoothly change the average batch size from $1$ to $n$. 
Next we show how to take the cache/disk IO effect into consideration and derive the ``optimal'' $p$ in theory, while
in the experimental part we show that the optimal average batch size can be large, depending on the problem and data.

To derive the optimal average batch size that yields minimal running time, we assume the computer needs 
$T_{\text{seq}}$ time to sequentially access $n$ samples, and $T_{\text{rand}}$ to randomly access the same number of samples. 
Therefore, when $|\mathcal{B}|=1$, each update costs within $T_{\text{rand}}/n$ time units. We call the ratio $T_{\text{seq}}/T_{\text{rand}}$ as the {\it cache effect ratio}.
\begin{corollary}\label{Coro:prac_B}
(\textbf{Optimal batch size with cache/disk IO effect}) If $T_{\text{seq}}/T_{\text{rand}}=\eta$, $\eta<1$, then the optimal average batch size will satisfy
the following equations:
\begin{equation}\label{optimal_B_equation}
\begin{aligned}
\mathbb{E}|\mathcal{B^*}|\approx(\frac{1}{\eta}-1)\frac{\xi - 1}{2 - \xi}, \quad \alpha=\frac{4\kappa}{\sqrt{\tau}n},\quad\\ \xi=\frac{\alpha^2\mathbb{E}|\mathcal{B^*}|^{2}}{1+\alpha^2\mathbb{E}|\mathcal{B^*}|^{2}-\sqrt{1+\alpha^2\mathbb{E}|\mathcal{B^*}|^{2}}}.
\end{aligned}
\end{equation}
\end{corollary}
Note that $\xi$ is also determined by $\mathbb{E}\mathcal{B}^*$ which makes the closed form solution intractable. {However, if we know condition number $\kappa$ and cache effect ratio $\eta$, the optimal batch size can be computed numerically}. To gain more insights, we plot $\frac{\xi-1}{2-\xi}$ as a function of $\alpha\cdot\mathbb{E}\mathcal{B}^*$ in Figure \ref{fig:optimal_B}(a), which shows the connection between the best batch size $\mathbb{E}\mathcal{B}^*$ and $\kappa$: 
in the well-conditioned regime we can select a larger batch size and in the ill-conditioned case a smaller batch size is better. Furthermore, Figure \ref{fig:optimal_B}(b) reveals the optimal average batch size changes with condition number and cache effect ratio $\eta$: 
Conceptually, if $\eta$ is smaller (sequential accesses are much faster than random accesses), then we are expected to do the full gradient update more frequently.

Our algorithm looks similar to SVRG---sometimes do a full gradient update while other times
select a single instance. However, we use a different book-keeping strategy---SVRG does not update the gradient snapshot and control variate (defined in~\eqref{gradient}) in between the two outer iterations, while SAGA++ will keep updating them even when batch size $=1$. 
Since we always keep the latest information, the convergence speed is always better than SVRG, and we leave the detailed discussion to Section~\ref{sec:onestep}.


\subsection{One step analysis: comparing Algorithm~\ref{alg:batch_method} with SVRG-style update}
\label{sec:onestep}
By far, we have only discussed the convergence speed under SAGA-style framework. In this section, we analyze why Algorithm~\ref{alg:batch_method} has faster convergence rate compared 
to the SVRG-style updates. This explains why SAGA++ (a special case of SVRG) is faster than SVRG, since they have the same data access pattern
and differs in update rules. 
Here SVRG-style means we do not update the control variate in~\eqref{gradient} before a new gradient snapshot is calculated, while in Algorithm~\ref{alg:batch_method} we store and update each gradient memorization $f'_i(\phi_i^t)$ as well as its summation once ``fresher'' gradient is available. 
Since the proposed framework in Algorithm~\ref{alg:batch_method} includes SAGA and SAGA++ as 
special cases, we call it ``SAGA-style'' update hereafter. 

The main advantage of SVRG-style update is that it needs less memory, however, since many machine learning problems can be formulated as
generalized linear model (GLM): $f_i(w)=f(w^\intercal x_i)$, so the gradient $f'_i(w)=f'(w^\intercal x_i)x_i$ is purely determined by $w^\intercal x_i$, and SAGA-style update need only to store
this scalar instead of the gradient vector for each sample. 
Therefore, for GLM problems the memory overhead of SAGA-style algorithm is simply an $\mathcal{O}(n)$ vector.

In terms of convergence rate, the following theory indicates that SAGA-style updates can better control the variance of gradient. First of all, we 
extend \eqref{gradient} to a more general variance reduced gradient defined by  $G(w^t)=f_i'(w^t)-g_i$, where $g_i=\alpha_i-\frac{1}{n}\sum_{j=1}^n\alpha_i$ can be any zero-mean control variate. The update rules for SVRG, SAGA, and SAGA++ can be written as $\alpha_i=f_i'(w^{\tau})$, where: 
    \begin{align}\nonumber
    \textbf{SVRG: }  \tau=kT\quad\\
    \textbf{SAGA: } 0\le \tau\le t\quad \\\nonumber
    \textbf{SAGA++: }  kT\le \tau \le t,
    \end{align}
    \label{eq:update_rules}
where $f_i'(w^\tau)=f_i'(\phi_i^t)$ is stored in memory, $T$ is the number of inner iterations inside each outer iteration and suppose the program have just finished the $k$-th outer iteration. We only consider the $|\mathcal{B}|=1$ case since we want to focus on the control variate rather than batch size. 
For each $i$, 
by regarding $\tau$ as a random variable, we can calculate its probability distribution as follows:
\begin{equation}\label{eq:distribution}
\begin{aligned}
\tau^{\text{SVRG}}=kT,\quad \tau^{\text{SAGA}}=
\begin{cases}
0, & p=(1-\frac{1}{n})^t\\
1, & p=\frac{1}{n}(1-\frac{1}{n})^{t-1}\\
\vdots &\\
t, & p=\frac{1}{n}
\end{cases}
,\\
\tau^{\text{SAGA++}}=
\begin{cases}
kT, & p=(1-\frac{1}{n})^{t-kT}\\
kT+1, & p=\frac{1}{n}(1-\frac{1}{n})^{t-kT-1}\\
\vdots &\\
t, & p=\frac{1}{n}.
\end{cases}
\end{aligned}
\end{equation}
To see the difference of convergence rate between those update rules, we introduce the following lemmas:
\begin{lemma}
If we use the distance $\Vert w^t-w^*\Vert^2$ as a metric to the sub-optimality, then we have:
\begin{equation}
\label{eq:distance-improvement}
\begin{aligned}
\mathbb{E}[\Vert w^{t+1}-w^* \Vert^2|\mathcal{F}_t] \le (1-\gamma\mu)\Vert w^t-w^* \Vert^2\\
+(4L\gamma^2-\frac{2\mu\gamma}{L}-2\mu\gamma^2)f^{\delta}(w^t)\\
+2\gamma^2\mathbb{E}[\Vert \alpha_i -f_i'(w^*) \Vert^2|\mathcal{F}_t], 
\end{aligned}
\end{equation}
where the expectation is taken over the choices of $i$~($\mathcal{F}_t$ is the $\sigma$-algebra at time $t$), $f^{\delta}(w)=f(w)-f(w^*)-f'(w^*)(w-w^*)$ is the Bregman divergence and we have $0\le f^{\delta}(w)\le F(w)-F(w^*)$.
\end{lemma}
The first two terms in~\eqref{eq:distance-improvement} is related to the distance between the current and optimal solution, 
only the last term involves the control variate $\alpha_i$ in different update rules, that is exactly what we are interested in, which
can be further bounded by: 
\begin{lemma}\label{le:grad_diff_bound}
For an algorithm with $P(\tau=j)=p_j$, we can upper bound the gradient difference term:
\begin{equation}\label{eq:control-variance}
\mathbb{E}[\Vert \alpha_i-f_i'(w^*)\Vert^2|\mathcal{F}_0] \le 2L\sum_{j=1}^t p_j F^{\text{sub}}(w^j). 
\end{equation}
\end{lemma}
 To see the change of $F^{\text{sub}}(w^t)=F(w^t)-F(w^*)$ with $t$,  we claim that those variance reduction methods is expected to decrease in each step as long as $\gamma$ is small(but keep to some constant):
\begin{proposition}\label{co:improve}
For strongly convex function $f(w)$, define the update rule $w^{t+1}=w^t-\gamma G(w^t)$ (we ignore the regularization term for simplicity).
If we want the function value to be a super-martingale, i.e. $\mathbb{E}[f(w^{t+1})|\mathcal{F}_t]\le f(w^t)$, then for SGD we require $\gamma\to 0$. But for variance reduction methods, since the variance of $G(w^t)$ goes to zero as fast as $f(w^t)-f^*$ (see appendix for details), a constant step size is enough.
\end{proposition}

Finally we can compare the update rules listed in ~\eqref{eq:update_rules} by Proposition \ref{co:improve}: 
Since the upper bound of distance improvement in \eqref{eq:distance-improvement} is determined by the variation of control variate $\mathbb{E}[\Vert \alpha_i-f'_i(w^*) \Vert^2|\mathcal{F}_0]$ and it is further upper bounded by \eqref{eq:control-variance}, this can be seen as a weighted sum of expected function suboptimal $\mathbb{E}[F^{\text{sub}}|\mathcal{F}_0]$ and further from Corollary \ref{co:improve} we know $F^{\text{sub}}$ is expected to decrease at each iteration, so we can conclude that
more ``weight'' $p_j$ should be put on smaller $\mathbb{E}[F^{\text{sub}}|\mathcal{F}_0]$, in another word, a good update rule should keep all the stochastic gradient active, rather than stale for too long. Therefore,  by \eqref{eq:distribution} we can observe that the distribution of $p_j$ in SAGA++ is strictly better than both 
SAGA and SVRG, which indicates that SAGA++ has a faster convergence rate while keeping the same computational cost. 
\subsection{Lazy update for $\ell_1$ regularization\label{sec:lazy}}

\begin{figure}[!th]

\centering

\begin{tikzpicture}[thick,scale=0.6, every node/.style={transform shape}]
\tikzset{cross/.style={cross out, draw=black, minimum size=2*(#1-\pgflinewidth), inner sep=0pt, outer sep=0pt},
cross/.default={1pt}}

\draw[thick] (0, 0) rectangle (1,5);
\draw[thick,cyan,fill=cyan] (0.5, 0.5) circle [radius=0.3];
\draw[thick,dashed] (0.5, 1.0) -- (0.5, 2);
\draw[thick,cyan,fill=cyan] (0.5, 2.5) circle [radius=0.3];
\draw[thick,dashed] (0.5, 3.0) -- (0.5, 4.8);
\draw[->,thick] (1.3, 0.5) -- (2.7, 0.5);
\draw[->,thick] (1.3, 2.5) -- (2.7, 2.5);
\node[above] at (0.5, 5) {$t=0$};

\draw[thick] (3, 0) rectangle (4,5);
\draw[thick,cyan] (3.5, 0.5) circle [radius=0.3];
\draw[thick,dashed] (3.5, 1.0) -- (3.5, 2);
\draw[thick,cyan,fill=cyan] (3.5, 2.5) circle [radius=0.3];
\draw[thick,dashed] (3.5, 3.0) -- (3.5, 4.8);
\draw[->,thick] (4.3, 0.5) -- (5.7, 0.5);
\draw[->,thick] (4.3, 2.5) -- (5.7, 2.5);
\draw[thick] (5,.5) node[cross=4pt,red] {};
\node[above] at (5, .5) [scale=0.74]{$\mathsf{prox}(w^{(1)}-c)$};
\node[above] at (3.5, 5) {$t=1$};

\draw[thick] (6, 0) rectangle (7,5);
\draw[thick,cyan] (6.5, 0.5) circle [radius=0.3];
\draw[thick,dashed] (6.5, 1.0) -- (6.5, 2);
\draw[thick,cyan,fill=cyan] (6.5, 2.5) circle [radius=0.3];
\draw[thick,dashed] (6.5, 3.0) -- (6.5, 4.8);
\draw[->,thick] (7.3, 0.5) -- (8.7, 0.5);
\draw[->,thick] (7.3, 2.5) -- (8.7, 2.5);
\draw[thick] (8,.5) node[cross=4pt,red] {};
\node[above] at (8, .5) [scale=0.74]{$\mathsf{prox}(w^{(2)}-c)$};
\node[above] at (6.5, 5) {$t=2$};

\draw[thick] (9, 0) rectangle (10,5);
\draw[thick,cyan,fill=cyan] (9.5, 0.5) circle [radius=0.3];
\draw[thick,dashed] (9.5, 1.0) -- (9.5, 2);
\draw[thick,cyan,fill=cyan] (9.5, 2.5) circle [radius=0.3];
\draw[thick,dashed] (9.5, 3.0) -- (9.5, 4.8);
\node[above] at (9.5, 5) {$t=3$};

\node[left] at (0, 2.5) {$i$};
\node[left] at (0, 0.5) {$j$};
\draw[->,ultra thick, dashed] (3.5, 0) arc (240:300:6.05);
\node[thick, above] at (6.5, -.85) {recover};
\node[below] at (6.5, -.8) [scale=1.2]{$\mathsf{prox}\big(\mathsf{prox}(w^{(1)}-c)-c\big)$};
\end{tikzpicture}
\caption{\label{fig:lazy_update} The illustration of lazy-update technique. 
We count the proximal operations that have been delayed (in this figure there are two) and recover it at once.
}
\end{figure}
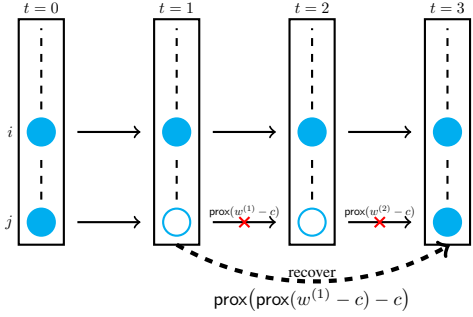

For sparse datasets and $f_i(w)=f(w^\intercal x_i)$, the stochastic gradient $f'_i(w^\intercal x_i)x_i$ has the same indices of zero elements as $x_i$. 
However, the $\bar{u}$ vector in update rule~\eqref{gradient} is a dense vector, which will lead to updating all the $d$ variables. 
To reduce the time complexity back to $O(\texttt{nnz}(x_i))$ per step, a ``lazy update'' technique was discussed in \cite{schmidt2013minimizing} for $\ell_2$ regularization. 
The main idea is that instead of performing an immediate update to all the variables, we only update variables associate with nonzero elements
in $x_i$. In the following, we derive the lazy update technique for $\ell_1$ regularization. 
As an illustration, Figure \ref{fig:lazy_update} shows a simple case where index $j$ has two consecutive zero elements in data vectors chosen at time $t=1$ and $t=2$, or $x_{i_1}[j]=x_{i_2}[j]=0$. So the updates of $w^{(2)}[j]$ and $w^{(3)}[j]$ are:
$$
\begin{aligned}
w^{(2)}[j] &= \mathsf{prox}\big( w^{(1)}[j] - c\big)\\
w^{(3)}[j] &= \mathsf{prox}\Big(\mathsf{prox}\big(w^{(1)}[j]-c\big)-c\Big),
\end{aligned}
$$
where $c=\frac{1}{n}\sum_{i=1}^nf'_{i}(x_{i}^\intercal \phi_{i}^{(1)})x_{i_1}[j]$. Now it remains to calculate the nested proximal operations, which can be effectively calculated by following theorem:
\begin{theorem}\label{th:seq_prox}
Let $g(x)=\eta|x|$, for all $x, c\in \mathbb{R}$ we have:
$
\underbrace{\mathsf{prox}_g\Big(\mathsf{prox}_g\big(\cdots\mathsf{prox}_g}_\text{n \text{operations}}(x-c)\cdots-c\big)-c\Big)=P_g(x, \eta, c, n).
$
Where $P_g$ is a simple, piecewise linear function.
\end{theorem}
Due to the space limit, we left the detailed formulation of $P_g$ in appendix. 
Upon finishing this paper, we found lazy update for $\ell_1$ regularization has also been discussed recently in~\cite{konevcny2016mini}. However, 
we still include our formal proof here for the completeness of this paper. 
\subsection{Extension: parallel computing scenario}
The fact that doing one step full gradient update is faster than $n$-step stochastic gradient update dues not only to cache/IO read; similar idea can also be applied to a variety of parallel optimization algorithms, in a more implicit way: when doing full gradient descent, it is trivial to make use of multiprocessing to speedup our program. In contrast, for mini-batch stochastic gradient upgrade when batch size is much smaller than the available CPU cores, many of the computing resources are
idle. Although many first order methods have their asynchronous versions that alleviate this problem to some degree~\cite{recht2011hogwild,leblond2016asaga,reddi2015variance,hsieh2015passcode}, because of the inconsistent paces between workers, these algorithms are suboptimal. So if we come back to synchronous updates and given that only full gradient calculation can be significantly accelerated, then the same quantity $T_{\text{seq}}/T_{\text{rand}}$ becomes a deterministic factor that affects how frequently one should perform full batch update. 

\section{Experimental Results}\label{sec:experiment}
\begin{table*}[h]
\caption{Dataset statistics. Time for sequential accessing $n$ samples is measured by one computation of full gradient $f'(w)$, while time for random accesses is measured by computing $n$ random gradient components $f_i'(w)$. \label{tab:datasets}}
\centering
\begin{tabular}{ccccccc}
\toprule
\multirow{2}[2]{*}{Dataset\textsuperscript{$\dagger$}} & \multirow{2}[2]{*}{Size(GB)} & \multirow{2}[2]{*}{\#Sample} & \multirow{2}[2]{*}{\#Feature} & \multirow{2}[2]{*}{nnz ratio} & \multicolumn{2}{c}{Time to access whole data(sec)}\\
\cmidrule(lr){6-7}
& &  &  &   & Sequential  & Random \\
\midrule
kddb & 5.13 &19,264,097 &  29,890,095 & 9.84e-7 & 3.91 & 11.43 \\
avazu & 5.04 &25,832,830 & 999,962 & 1.50e-5 & 4.14 & 9.08 \\
criteo & 26.74 &45,840,617 & 999,999 & 3.90e-5 & 14.07 & 30.51\\
\bottomrule
\multicolumn{7}{l}{\textsuperscript{$\dagger$}Download from \url{https://www.csie.ntu.edu.tw/~cjlin/libsvmtools/datasets/}}
\end{tabular}
\end{table*}\vspace*{-5pt}
\begin{figure*}[h]
    \centering
    \includegraphics[width=\linewidth]{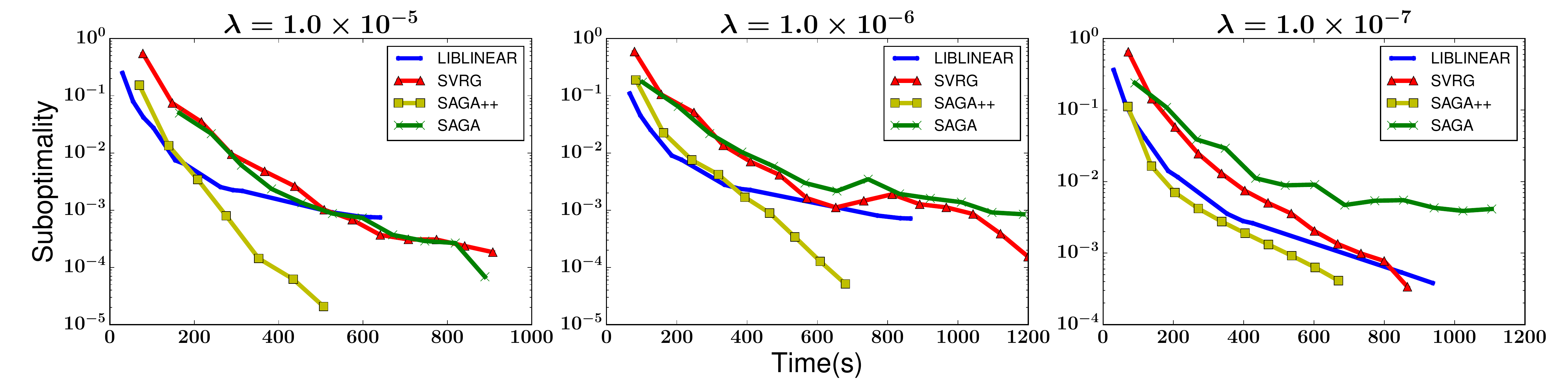}\vspace*{-5pt}
    \caption{Running time comparison on kddb dataset with different regularization parameters ($\lambda$). Result shows that our SAGA++ algorithm is faster than competitors under different regularization parameters. 
    }
    \label{fig:bsaga_exp}
\end{figure*}\vspace*{-3pt}
\begin{figure*}[h!]
    \centering
    \includegraphics[width=\linewidth]{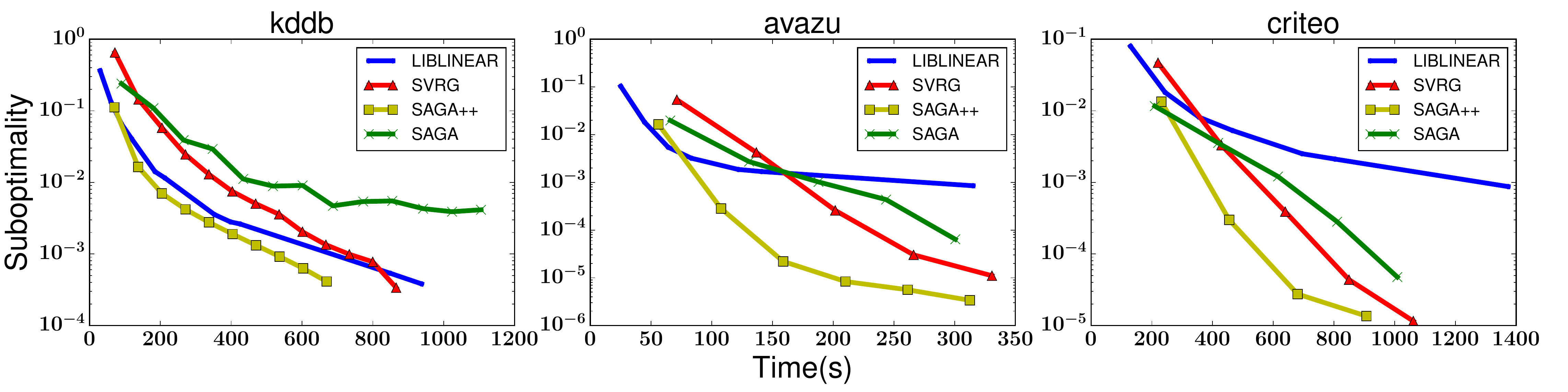}\vspace*{-5pt}
    \caption{Running time comparison among different data ($\lambda=1.0\times 10^{-7}$ for all data). Meta-information can be found in Table~\ref{tab:datasets}.}
    \label{fig:my_label}
\end{figure*}

We compare SAGA++ with SAGA~\cite{defazio2014new}, SVRG~\cite{johnson2013accelerating} and LIBLINEAR~\cite{fan2008liblinear} (proximal Newton method) on solving the $\ell_1$-regularized
logistic regression: 
\begin{equation}
w^*=\argmin_w \frac{1}{n}\sum_{i=1}^n\log\big(1+\exp(y_ix_i^\intercal w)\big) + \lambda\Vert w \Vert_1,
\end{equation}
where $(x_i, y_i)$ are feature-response pairs. To make a fair comparison, all the algorithms are implemented based on the LIBLINEAR code base, and we tried to optimize each algorithm. For each outer iteration in SVRG/SAGA++ we choose $m=1.5n$ inner iterations, this amounts to $\mathbb{E}|\mathcal{B}|=1.67$ and close to the experiments in \cite{johnson2013accelerating} where $m=2n$.
The lazy update for $\ell_1$ regularization is also implemented for all the variance reduced methods. All the datasets can be downloaded from LIBSVM website. 

First, we compare all the algorithms on kddb dataset with different regularization parameters. The results in Figure~\ref{fig:bsaga_exp} shows that
SAGA++ outperforms other algorithms for all the three choices of parameters. Indeed, $\lambda=10^{-6}$ (the middle figure) is the best parameter in terms
of prediction accuracy, so our comparison covers both large and small $\lambda$s. 

Next, we compare the running time of  all the algorithms on three datasets in Figure~\ref{fig:my_label}. 
The results show that SAGA++ is faster than all the competitors on these three datasets. 
We conclude our experimental results by the following observations: 
(1) Although SAGA has faster convergence in terms of ``number of data access'' (see Figure~\ref{fig:optimal_B}-c), 
SVRG often outperforms SAGA due to faster sequential access. Our algorithm, SAGA++, has a sequential access stage like SVRG, while uses the most up-to-date gradient information
at the random update stage, thus outperforms both SVRG and SAGA in all cases. 
(2) The lazy update (briefly discussed in Section~\ref{sec:lazy}) accelerates SAGA/SVRG/SAGA++ a lot; without such technique all the variance reduction methods will be much slower
than LIBLINEAR. However, with such technique they can outperform the LIBLINEAR implentation of proximal Newton methods.

\section{Conclusions and Discussions}   We study the unified framework for variance reduction methods with stochastic batch size and prove the linear convergence rate in strongly convex finite sum functions with continuous regularizer. 
    We show that choosing batch size always equals to $1$ (equivalent to SAGA) leads to the best rate in terms of number of data accesses; 
    however, it is not optimal in terms of running time, so we develop a new SAGA++ algorithm. 
    We demonstrate that SAGA++ outperforms SAGA and SVRG in terms of running time, both in theory and in practice. 
   
   One reason that SAGA++ outperforms other VR methods is due to the cache/IO effect, although we only shows in-memory optimization,
   one can imagine that when data is too large to fit in memory and an out-of-core solver is needed, the overhead of IO will be even more significant. 
   In this setting we would expect an even more greater advantage over SVRG/SAGA. Another important reason is that SAGA++ update its control variate more frequently, making the stochastic gradient a lower variance estimator, so intuitively its more close to gradient descent.
   
\section*{Acknowledgements}
The authors acknowledge the support of NSF via IIS-1719097 and the computing resources provided by Google cloud and Nvidia.

\bibliography{ref}
\bibliographystyle{icml2018}

\appendix
\renewcommand{\theequation}{A\arabic{equation}}
\onecolumn
\section{Appendix}
\subsection{Proof of Lemma~1}
It is straight forward to see:
$$
\begin{aligned}
\mathbb{E}\bar{H}_{t+1}=\mathbb{E}[\frac{1}{n}\sum_{i=1}^n f^{\delta}_i(\phi_{i}^{t+1})]
&=\mathbb{E}[\frac{1}{n}\big(\sum_{i\in\mathcal{B}}f^{\delta}_i(w^t)+\sum_{i\notin\mathcal{B}}f^{\delta}_i(\phi_i^t)\big)]\\
&=\mathbb{E}[\mathbb{E}[\frac{1}{n}\big(\sum_{i\in\mathcal{B}}f^{\delta}_i(w^t)+\sum_{i\notin\mathcal{B}}f^{\delta}_i(\phi_i^t)\big)|\mathcal{B}]|\quad|\mathcal{B}|=B]\\
&=
\frac{1}{n}
\big(
\frac{\mathbb{E}|\mathcal{B}|}{n}
\sum_{i=1}^nf_i^{\delta}(w^t)+
\frac{n-\mathbb{E}|\mathcal{B}|}{n}
\sum_{i=1}^n
f_i^{\delta}(\phi_i^t)
\big)\\
&=\frac{\mathbb{E}|\mathcal{B}|}{n}f^{\delta}(w^t)+\frac{n-\mathbb{E}|\mathcal{B}|}{n}\bar{H}_t
\end{aligned}
$$
The second line of equality comes from the rule of total expectation, where the inner expectation is taken over the index set $\mathcal{B}$, and the outer expectation is taken over the set cardinality $|\mathcal{B}|$.
\subsection{Proof of Lemma~2}
The proof technique is similar to SAGA, as well as a useful inequality (Lemma 4 in~\cite{defazio2014saga}):
\begin{equation}\label{bound_defazio}
\begin{aligned}
    & f(x)\ge f(y)+\langle f'(y),x-y \rangle+\frac{1}{2(L-\mu)}\Vert f'(x)-f'(y) \Vert^2\\
&\qquad + \frac{\mu L}{2(L - \mu)}\Vert  x-y\Vert^2 -\frac{\mu}{(L-\mu)}\langle f'(x)-f'(y), x-y \rangle.
\end{aligned}
\end{equation}
First of all, by the update rule~(2):
\begin{equation}\label{distance_suboptimal}
\begin{aligned}
\Vert w^{t+1}-w^* \Vert^2&=\Vert \mathsf{Prox}_{\gamma g}(w^t-\gamma G(w^t))-\mathsf{Prox}_{\gamma g}(w^*-\gamma f'(w^*)) \Vert^2\\
&\le \Vert w^t-\gamma G(w^t)-w^*+\gamma f'(w^*) \Vert^2 \\
&= \Vert w^t-w^* \Vert^2 - 2\gamma \langle w^t-w^*, G(w^t)-f'(w^*) \rangle + \gamma^2\Vert G(w^t)-f'(w^*) \Vert^2.
\end{aligned}
\end{equation}
The inequality follows from non-expansiveness of proximal operator, notice that our stochastic gradient $G(w^t)$ is unbiased, take the expectation to the second term and apply \eqref{bound_defazio} to each $f_i$ and the average over all $i$ will goes to:
\begin{equation}\label{inner_bound_part}
\begin{aligned}
-\mathbb{E}[\langle w^t-w^*, G(w^t)-f'(w^*) \rangle]=-\langle w^t-w^*, f'(w^t)-f'(w^*)\rangle \\
\le \langle w^t-w^*, f'(w^*) \rangle + \frac{L-\mu}{L}[f(w^*)-f(w^t)]-\frac{\mu}{2}\Vert w^*-w^t \Vert^2\\
-\frac{1}{2Ln}\sum_{i=1}^n\Vert f_i'(w^*)-f_i'(w^t) \Vert^2-\frac{\mu}{L}\langle f'(w^*), w^t-w^* \rangle.
\end{aligned}
\end{equation}
Next we bound the last term in \eqref{distance_suboptimal}:
\begin{equation}
\begin{aligned}
\mathbb{E}\left\Vert f'(w^*) - G(w^t) \right\Vert^2
&=\mathbb{E}\left\Vert f'(w^*) - \frac{1}{|\mathcal{B}|}\sum_{i\in\mathcal{B}}f_i'(w)+\frac{1}{|\mathcal{B}|}\sum_{i\in\mathcal{B}}f_i'(\phi_i^t)-\frac{1}{n}\sum_{i=1}^n f_i'(\phi_i^t) \right\Vert^2\\
&=\mathbb{E}\Bigg\Vert \Big[\frac{1}{n}\sum_{i=1}^nf'_i(w^t)-\frac{1}{|\mathcal{B}|}\sum_{i\in\mathcal{B}}f'_i(w^t)-f'(w^*)+\frac{1}{|\mathcal{B}|}\sum_{i\in\mathcal{B}}f'_i(w^*)\Big] \\
   &\quad -\Big[\frac{1}{n}\sum_{i=1}^nf'_i(\phi_i^t)-\frac{1}{|\mathcal{B}|}f'_i(\phi_i^t)-f'(w^*)+\frac{1}{|\mathcal{B}|}\sum_{i\in\mathcal{B}}f'_i(w^*)\Big] \\
   &\quad +f'(w^*)-\frac{1}{n}\sum_{i=1}^nf'_i(w^t) \Bigg\Vert^2\\
&\overset{*}{=} \mathbb{E}\Bigg\Vert \Big[\frac{1}{n}\sum_{i=1}^nf'_i(w^t)-\frac{1}{|\mathcal{B}|}\sum_{i\in\mathcal{B}}f'_i(w^t)-f'(w^*)+\frac{1}{|\mathcal{B}|}\sum_{i\in\mathcal{B}}f'_i(w^*)\Big] \\
   &\quad -\Big[\frac{1}{n}\sum_{i=1}^nf'_i(\phi_i^t)-\frac{1}{|\mathcal{B}|}\sum_{i\in\mathcal{B}}f'_i(\phi_i^t)-f'(w^*)+\frac{1}{|\mathcal{B}|}\sum_{i\in\mathcal{B}}f'_i(w^*)\Big]\Bigg\Vert^2 \\
   &\quad + \Big \Vert f'(w^*)-\frac{1}{n}\sum_{i=1}^n f'_i(w^t) \Big\Vert^2.
\end{aligned}
\end{equation}
In equation $\overset{*}{=}$ we use the property that $\mathbb{E}[X^2]=\mathbb{E}[X-\mathbb{E}[X]]^2+\mathbb{E}[X]^2$, now use the inequality $\Vert X+Y \Vert^2\le (1+\beta)\Vert X \Vert^2+(1+\beta^{-1})\Vert Y \Vert^2$, $\beta>0$ to the first term:

\begin{equation}\label{variance_bound}
\begin{aligned}
    \mathbb{E}\left\Vert f'(w^*) - G(w^t) \right\Vert^2 &\le
    (1+\beta)\mathbb{E}\Bigg\Vert \frac{1}{n}\sum_{i=1}^nf'_i(w^t)-\frac{1}{|\mathcal{B}|}\sum_{i\in\mathcal{B}}f'_i(w^t)-f'(w^*)+\frac{1}{|\mathcal{B}|}\sum_{i\in\mathcal{B}}f'_i(w^*) \Bigg\Vert^2 \\
    &\quad +(1+\beta^{-1})\mathbb{E}\Bigg\Vert \frac{1}{n}\sum_{i=1}^nf'_i(\phi_i^t)-\frac{1}{|\mathcal{B}|}\sum_{i\in\mathcal{B}}f'_i(\phi_i^t)-f'(w^*)+\frac{1}{|\mathcal{B}|}\sum_{i\in\mathcal{B}}f'_i(w^*) \Bigg\Vert^2\\
    &\quad +\beta\cdot\Big \Vert f'(w^*)-\frac{1}{n}\sum_{i=1}^n f'_i(w^t) \Big\Vert^2.
\end{aligned}
\end{equation}
Next we bound the first and second terms again by variance decomposition, for simplicity we only take the first term as example:
\begin{equation}\label{variance_bound_part1}
\begin{aligned}
&\mathbb{E}\Bigg\Vert \frac{1}{n}\sum_{i=1}^nf'_i(w^t)-\frac{1}{|\mathcal{B}|}\sum_{i\in\mathcal{B}}f'_i(w^t)-f'(w^*)+\frac{1}{|\mathcal{B}|}\sum_{i\in\mathcal{B}}f'_i(w^*) \Bigg\Vert^2\\
=&\mathbb{E}\Bigg \Vert \frac{1}{|\mathcal{B}|}\sum_{i\in\mathcal{B}}\Big (f_i'(w^*)-f_i'(w^t)\Big) \Bigg\Vert^2- \Bigg\Vert \frac{1}{n}\sum_{i=1}^n \Big(f'_i(w^*)-f'_i(w^t)\Big)\Bigg\Vert^2\\
\overset{(1)}{\le} & \mathbb{E}\Big(\frac{1}{|\mathcal{B}|}\sum_{i\in\mathcal{B}}\big\Vert f_i'(w^*)-f_i'(w^t) \big\Vert^2\Big)-
\Bigg\Vert \frac{1}{n}\sum_{i=1}^n \Big(f'_i(w^*)-f'_i(w^t)\Big)\Bigg\Vert^2\\
=&\frac{1}{n}\sum_{i=1}^n\Vert f_i'(w^*)-f_i'(w^t) \Vert^2 - \Bigg\Vert \frac{1}{n}\sum_{i=1}^n \Big(f'_i(w^*)-f'_i(w^t)\Big)\Bigg\Vert^2\\
\overset{(2)}{\le} &\frac{1}{n}\sum_{i=1}^n\Vert f_i'(w^*)-f_i'(w^t) \Vert^2,
\end{aligned}
\end{equation}
$\overset{(1)}{\le}$ is by RMS-AM inequality, and in $\overset{(2)}{\le}$ we drop the negative term. Similarly,
$$
\mathbb{E}\Bigg\Vert \frac{1}{n}\sum_{i=1}^nf'_i(\phi_i^t)-\frac{1}{|\mathcal{B}|}\sum_{i\in\mathcal{B}}f'_i(\phi_i^t)-f'(w^*)+\frac{1}{|\mathcal{B}|}\sum_{i\in\mathcal{B}}f'_i(w^*) \Bigg\Vert^2\le \frac{1}{n}\sum_{i=1}^n\Vert f_i'(w^*)-f_i'(\phi_i^t) \Vert^2.
$$
Plug \eqref{variance_bound_part1} into \eqref{variance_bound} we get:
\begin{equation}\label{variance_bound_final}
\begin{aligned}
\mathbb{E}\left\Vert f'(w^*) - G(w^t) \right\Vert^2 
&\le \frac{(1+\beta)}{n}\sum_{i=1}^n\Vert f_i'(w^*)-f_i'(w^t) \Vert^2 + \frac{(1+\beta^{-1})}{n}\sum_{i=1}^n\Vert f_i'(w^*)-f_i'(\phi_i^t) \Vert^2\\
&\quad - \beta\Vert f'(w^t)-f'(w^*)\Vert^2.
\end{aligned}
\end{equation}
Combining \eqref{distance_suboptimal},\eqref{inner_bound_part},\eqref{variance_bound_final} becomes (5) immediately:
$$
\begin{aligned}
\Vert w^{t}-w^*\Vert^2-\mathbb{E}\Vert w^{t+1}-w^*\Vert^2 \ge &\gamma\mu\Vert w^t-w^* \Vert^2
- (2\gamma^2-\gamma/L)\mathbb{E}\Vert f_i'(w^t)-f'_i(w^*) \Vert^2 \\\nonumber
&+ \gamma^2\Vert f'(w^t)-f'(w^*) \Vert^2+\frac{2\gamma(L-\mu)}{L}f^{\delta}(w^t)-4\gamma^2 L\bar{H}_t. 
\end{aligned}
$$
\subsection{Proof of Theorem~3}
It follows directly from Lemma~1 and 2:
\begin{equation}
\begin{aligned}
\mathcal{L}_t-\mathbb{E}\mathcal{L}_{t+1}&=c(\bar{H}_t - \mathbb{E}\bar{H}_{t+1})+(\Vert w^t-w^* \Vert^2 - \mathbb{E}\Vert w^{t+1}-w^* \Vert^2)\\
&\ge c\Big(\frac{\mathbb{E}|\mathcal{B}|}{n}-\frac{2(1+\beta^{-1})\gamma^2L}{c}\Big)\bar{H}_t+\gamma\mu\Vert w^t-w^* \Vert^2+(2\mu\beta\gamma^2+\frac{2\gamma(L-\mu)}{L}-\frac{c\cdot\mathbb{E}|\mathcal{B}|}{n})f^{\delta}(w^t)\\
&\qquad + (\frac{\gamma}{L}-(1+\beta)\gamma^2)\mathbb{E}\Vert f_i'(w^t)-f'(w^*) \Vert^2\\
&\overset{?}{\ge} c\Big(\frac{|\mathcal{B}|}{n}-\frac{2(1+\beta^{-1})\gamma^2L}{c}\Big)\bar{H}_t+\gamma\mu\Vert w^t-w^* \Vert^2\\
&\ge \rho\mathcal{L}_t,
\end{aligned}
\end{equation}
where $\rho=\min(\frac{|\mathcal{B}|}{n}-\frac{2(1+\beta^{-1})\gamma^2L}{c}, \gamma\mu)$, the last inequality $\overset{?}{\ge}$ comes with following condition:
\begin{equation}\label{eq:gamma_condition_1}
\begin{aligned}
2\mu\beta\gamma^2+\frac{2\gamma(L-\mu)}{L}-\frac{c|\mathcal{B}|}{n} &\ge 0\\ 
\frac{\gamma}{L}-(1+\beta)\gamma^2 &\ge 0,
\end{aligned}
\end{equation}
furthermore, to keep our algorithm moving forward, i.e. $\Vert w^t-w^* \Vert^2$ decreasing, we should also make sure such condition hold:
\begin{equation}\label{eq:gamma_condition_2}
\frac{|\mathcal{B}|}{n}-\frac{2(1+\beta^{-1})\gamma^2L}{c}\ge 0.
\end{equation}

\subsection{Proof of Proposition~1}
By plugging $\beta=2$, $c=\frac{n}{3L\mathbb{E}|\mathcal{B}|}$ into \eqref{eq:gamma_condition_1} it is easy to verify both inequalities hold.

\subsection{Proof of Proposition~2}
In this case we choose $\beta=1$. From Theorem~3 we know that with a suitable step size $\gamma$ and $c$, we have:
$$
\mathbb{E}\Vert w^t-w^*\Vert^2\le \mathbb{E}\mathcal{L}_t\le (1-\rho)^t\mathcal{L}_0=(1-\rho)^t\big[\Vert w^0-w^*\Vert^2+c\bar{H}_0\big].
$$
For the optimal convergence rate, we try to maximize the geometric factor $\rho=\min(\frac{\mathbb{E}|\mathcal{B}|}{n}-\frac{4\gamma^2L}{c}, \gamma\mu)$. Denote $\gamma_0$ as the solution of: $\frac{\mathbb{E}|\mathcal{B}|}{n}-\frac{4\gamma_0^2L}{c}= \gamma_0\mu$. Notice that $\rho(\gamma)=\gamma\mu$ is increasing with $\gamma$ when $\gamma \le \gamma_0=\frac{c}{8\kappa}\Big(\sqrt{1+\frac{16\kappa\mathbb{E}|\mathcal{B}|}{cn\mu}}-1\Big)$ and $\rho(\gamma)=\frac{\mathbb{E}|\mathcal{B}|}{n}-\frac{4\gamma^2L}{c}$ is decreasing when $\gamma>\gamma_0$. So the optimal step size should be $\gamma=\gamma_0$. However we should also verify that this step size indeed satisfies the condition in \eqref{eq:gamma_condition_1}. First of all:
\begin{equation}
\begin{aligned}
\gamma_0&=\frac{c}{8\kappa}\Big(\sqrt{1+\frac{16\kappa\mathbb{E}|\mathcal{B}|}{cn\mu}}-1\Big) \overset{(1)}{\le} \frac{c}{8\kappa}\sqrt{\frac{16\kappa\mathbb{E}|\mathcal{B}|}{cn\mu}}=\sqrt{\frac{c\mathbb{E}|\mathcal{B}|}{4nL}}\overset{(2)}{\le} \frac{1}{2L}.
\end{aligned}
\end{equation}
$\overset{(1)}{\le}$ comes from the fact that $\sqrt{1+x}-1\le \sqrt{x}$, $\overset{(2)}{\le}$ holds by choosing $c=\frac{\tau n}{L\mathbb{E}|\mathcal{B}|}$, where $\tau<1$ is a small constant. These two inequalities together ensure the upper bound part of \eqref{eq:gamma_condition_1}. As to the lower bound, we have $\sqrt{1+x}-1> \sqrt{x}-1$, so:
$$
\gamma_0>\frac{c}{8\kappa}\Big(\sqrt{\frac{16\kappa\mathbb{E}|\mathcal{B}|}{cn\mu}}-1\Big)\ge \frac{c\mathbb{E}|\mathcal{B}|L}{2n(L-\mu)} \Longrightarrow \tau \le \Big(\frac{1}{\frac{L}{L-\mu}+\frac{n}{4\kappa\mathbb{E}|\mathcal{B}|}}\Big)^2<1.
$$
So if we choose $\tau$ properly, both sides of \eqref{eq:gamma_condition_1} can be satisfied.

\subsection{Proof of Corollary 1, 2}
Following (7) we take a derivative to $\mathbb{E}|\mathcal{B}|$:
\begin{equation}
\frac{\partial f(\mathbb{E}|\mathcal{B}|)}{\partial \mathbb{E}|\mathcal{B}|}=\frac{(\alpha\mathbb{E}|\mathcal{B}|-1)^2\mathbb{E}|\mathcal{B}|}{\sqrt{1+\alpha^2\mathbb{E}|\mathcal{B}|^2}(\sqrt{1+\alpha^2\mathbb{E}|\mathcal{B}|^2}-1)^2}\ge 0,
\end{equation}
where $\alpha=\frac{4\kappa}{\sqrt{\tau}n}$, so there is no optimal batch size, and since we always want to access one data point, i.e. $|\mathcal{B}|\ge 1$ and SAGA style update is optimal.
\par
For Corollary 2, it is easy to see for our algorithm, which choose $|\mathcal{B}|=n$ with probability $p\ll 1$ and $|\mathcal{B}|=1$ with probability $1-p$, has average batch size $\mathbb{E}|\mathcal{B}|=np+1-p\approx np+1$. For each update, it takes on average time $\tau=n\eta\tau p+(1-p)\tau\approx (1+np\eta)\tau$. If we want to get a $\epsilon$-suboptimal solution, the total iteration will be $N=\frac{\log(\epsilon/\epsilon_0)}{\log(1-\rho)}\propto 1/\rho$, So the running time will be:
\begin{equation}
\begin{aligned}
T&\propto\frac{1+np\eta}{\sqrt{\frac{1}{\mathbb{E}|\mathcal{B}|^2}+\frac{16\kappa^2}{\tau n^2}}-\frac{1}{\mathbb{E}|\mathcal{B}|}}\\
&\approx \frac{(\mathbb{E}|\mathcal{B}|^2-\mathbb{E}|\mathcal{B}|)\eta + \mathbb{E}|\mathcal{B}|}{\sqrt{1+\alpha^2\mathbb{E}|\mathcal{B}|^2}-1}.
\end{aligned}
\end{equation}
For simplicity we denote $B=\mathbb{E}|\mathcal{B}|$. By taking the partial derivative and set it to zero $\partial T/\partial B=0$ can solve the best batch size:
\begin{equation}
\begin{aligned}
\big((2B|-1)\eta+1\big)(\sqrt{1+\alpha^2B^2}-1)=\big((B^2-B\big)\eta+B)\frac{\alpha^2B}{\sqrt{1+\alpha^2B^2}},
\end{aligned}
\end{equation}
after solving the above equation we get:
\begin{equation}
B=\left(\frac{1}{\eta}-1\right)\left(\frac{\xi-1}{2-\xi}\right), \quad \xi=\frac{\alpha^2B^2}{1+\alpha^2B^2-\sqrt{1+\alpha^2B^2}}.
\end{equation}
By showing the second order derivative $\partial^2T/\partial B^2\ge 0$ it's easy to verify that this solution is actually a global minimum.

\subsection{Proof of Lemma 4}
We begin with non-expansiveness of proximal operation:
\begin{equation}
\begin{aligned}
    \Vert w^{t+1}-w^* \Vert^2 &= \Vert\mathsf{Prox}_{\gamma g}(w^t-\gamma G(w^t))-\mathsf{Prox}_{\gamma g}(w^*-\gamma f'(w^*))\Vert^2 \\
    & \le \Vert w^t-\gamma G(w^t) - w^* + \gamma f'(w^*) \Vert^2\\
    &= \Vert w^t-w^*\Vert^2 - 2\gamma\langle w^t-w^*, G(w^t)- f'(w^*)\rangle+\gamma^2\Vert G(w^t)-f'(w^*) \Vert^2,
\end{aligned}
\end{equation}
where $f(w)=\frac{1}{n}\sum_{i=1}^nf_i(w)$ By taking expectation on each side and notice $G(w^t)$ is a unbiased estimation of $f'(w^t)$:
\begin{equation}
\begin{aligned}
\mathbb{E}\Vert w^{t+1}-w^* \Vert^2=\Vert w^t-w^* \Vert^2-2\gamma\langle w^t-w^*, f'(w^t)-f'(w^*)\rangle+\gamma^2\mathbb{E}\Vert G(w^t)-f'(w^*) \Vert^2,
\end{aligned}
\end{equation}
and then apply the following bounds for strongly convex function $f$:
\begin{equation}
\begin{aligned}
\langle w^t-w^*, f'(w^t)-f'(w^*)\rangle &\ge \mu\Vert w^t-w^*\Vert^2\\
\langle w^t-w^*, f'(w^t)-f'(w^*)\rangle &\ge \frac{1}{L}\Vert f'(w^t)-f'(w^*) \Vert^2,
\end{aligned}
\end{equation}
so the inner product term have a composite upper bound:
\begin{equation}
-2\gamma\langle w^t-w^*, f'(w^t)-f'(w^*)\rangle \le -\gamma(\mu\Vert w^t-w^* \Vert^2+\frac{1}{L}\Vert f'(w^t)-f'(w^*) \Vert^2)
\end{equation}
on the other hand, we can bound $\mathbb{E}\Vert G(w^t)-f'(w^*) \Vert^2$ as \eqref{variance_bound} but we only need to care about one sample in a batch case, since we are comparing SAGA with SVRG update style:
\begin{equation}
\mathbb{E}\Vert G(w^t)- f'(w^*) \Vert^2 \le 2\mathbb{E}\Vert f_i'(\phi_i^t)-f_i'(w^*) \Vert^2 + 2\mathbb{E}\Vert f_i'(w^t)-f_i'(w^*) \Vert^2-\Vert f'(w^t)- f'(w^*)\Vert^2.
\end{equation}
Remember we have proved above formula in \eqref{variance_bound_final}, for $\mathbb{E}\Vert f_i'(w^t)-f_i'(w^*) \Vert^2$ we have:
\begin{equation}
\begin{aligned}
\mathbb{E}\Vert f_i'(w^t)-f_i'(w^*) \Vert^2 
&\le \frac{2L}{n}\sum_{i=1}^n f_i(w^t)-f_i(w^*)-f_i'(w^*)^\intercal(w^t-w^*)\\
&=2L(f(w^t)-f(w^*)-f'(w^*)^\intercal(w^t-w^*)).
\end{aligned}
\end{equation}
Similarly, for $\Vert f'(w^t)-f'(w^*) \Vert^2$ we recall $f$ is a $\mu$-strongly convex function:
\begin{equation}
\Vert f'(w^t)-f'(w^*) \Vert^2 \ge 2\mu(f(w^t)-f(w^*)-f'(w^*)^\intercal(w^t-w^*)).
\end{equation}
Add those inequalities together:
\begin{equation}
\mathbb{E}\Vert w^{t+1}-w^* \Vert^2 \le (1-\gamma\mu)\Vert w^t-w^* \Vert^2+(4L\gamma^2-\frac{2\mu\gamma}{L}-2\mu\gamma^2)f^{\delta}(w^t)+2\gamma^2\mathbb{E}\Vert f_i'(\phi_i^t)-f_i'(w^*) \Vert^2.
\end{equation}

\subsection{Proof of Lemma 5}
Since we know the distribution of random variable $\tau$, also denote $t_s$ as the index of the latest gradient snapshot so for SVRG/SAGA++ $t_s=kT$ where $k$ is the number of outer iteration and $T$ is the length of inner iteration, for SAGA $t_s=0$ so in either method we have $t_s\ge 0$ then by conditional expectation relationship:
\begin{equation}
\begin{aligned}
\mathbb{E}[\Vert\alpha_i-f_i'(w^*)\Vert^2|\mathcal{F}_0]
&\overset{(1)}{=}\frac{1}{n}\sum_{k=1}^n\mathbb{E}[\Vert \alpha_k-f_k'(w^*) \Vert^2|\mathcal{F}_{t_s}]\\
&\overset{(2)}{=}\frac{1}{n}\sum_{k=1}^n\sum_{l=t_s}^{t}p_l\Vert f_k'(w^l)-f_k'(w^*)\Vert^2\\
&=\sum_{l=t_s}^t p_l\frac{1}{n}\sum_{k=1}^n \Vert f_k'(w^l)-f_k'(w^*) \Vert^2\\
&\le 2L\sum_{l=t_s}^t p_l(f(w^l)-f(w^*)-f'(w^*)(w^l-w^*)),
\end{aligned}
\end{equation}
$\overset{(1)}{=}$ is taken over the choices of $i$, while $\overset{(2)}{=}$ is taken over the random variable $\tau$ in $\alpha_k=f'_k(w^\tau)$. Because the regularization function $g(w)$ is convex, and from optimal condition we know: $-f'(w^*)\in \partial g(w^*)$, we have:
\begin{equation}
\begin{aligned}
f(w^l)-f(w^*)-f'(w^*)(w^l-w^*)
&=f(w^l)-f(w^*)+v^l(w^l-w^*)\\
&\le f(w^l)-f(w^*)+g(w^l)-g(w^*)\\
&=F(w^l)-F(w^*),
\end{aligned}
\end{equation}
where $v^l\in \partial g(w^l)$. Finally we have $\mathbb{E}[\Vert\alpha_i-f_i'(w^*)\Vert^2|\mathcal{F}_0]\le 2L\sum_{l=t_s}^t p_l(F(w^l)-F(w^*))$.

\subsection{Proof of Proposition 3}
Recall the quadratic upper bound of $L$-Lipschitz function:
\begin{equation}
f(w^t-\gamma G(w^t))\le f(w^t)-\gamma \nabla f^\intercal (w^t)G(w^t)+\frac{L\gamma^2}{2}\Vert G(w^t)\Vert^2.
\end{equation}
By taking the expectation,
\begin{equation}
\begin{aligned}
\mathbb{E}[f(w^t-\gamma G(w^t))|\mathcal{F}_t] &\le f(w^t)-\gamma\Vert f(w^t)\Vert^2+\frac{L\gamma^2}{2}\mathbb{E}[\Vert G(w^t)\Vert^2|\mathcal{F}_t]\\
&\le f(w^t)-(\gamma-\frac{L\gamma^2}{2})\Vert \nabla f(w^t) \Vert^2 + \frac{L\gamma^2}{2}\mathsf{Var}[G(w^t)].
\end{aligned}
\end{equation}
On the other hand, for $\mu$-strongly convex $f$, we have:
\begin{equation}
\Vert\nabla f(w^t)\Vert^2\ge 2\mu(f(w^t)-f^*),
\end{equation}
so if $\mathsf{Var}[G(w^t)]$ also converges to zero at the order of $f^{\text{sub}}(w^t)=f(w^t)-f^*$ then $\gamma$ can keep to a small constant rather than damping like SGD. In fact \cite{xiao2014proximal}(Corollary 3) already proved it for SVRG, here we prove a similar result for SAGA style update:
\begin{equation}
\begin{aligned}
    \mathsf{Var}[G(w^t)|\mathcal{F}_s]&=\mathbb{E}\Big[\Big\Vert \nabla f_{i_k}(w^t)-\nabla f_{i_k}(\phi_{i_k}^t)-\frac{1}{n}\sum_{j=1}^n \big(\nabla f_j(w^t)-\nabla f_j(\phi_j^t)\big)\Big\Vert^2\Big|\Big.\mathcal{F}_s\Big]\\
    &= \mathbb{E}\Big[ \Big\Vert \nabla f_{i_k}(w^t)-\nabla f_{i_k}(\phi_{i_k}^t) \Big\Vert^2 \Big|\Big.\mathcal{F}_s\Big]-\Big\Vert \frac{1}{n}\sum_{j=1}^n \big(\nabla f_j(w^t)-\nabla f_j(\phi_j^t)\big)\Big\Vert^2\\
    &\le \mathbb{E}\Big[\mathbb{E}\Big[ \Big\Vert \nabla f_{i_k}(w^t)-\nabla f_{i_k}(\phi_{i_k}^t) \Big\Vert^2 \Big|\mathcal{F}_t\Big|\mathcal{F}_s\Big]\Big]\\
    &= \frac{2}{n}\sum_{j=1}^n\mathbb{E}\big[\big\Vert \nabla f_{j}(w^t)-\nabla f_{j}(w^*) \big\Vert^2|\mathcal{F}_s\big]+\frac{2}{n}\sum_{j=1}^n\mathbb{E}\big[\big\Vert \nabla f_{j}(\phi_j^t)-\nabla f_{j}(w^*) \big\Vert^2|\mathcal{F}_s\big]\\
    &\le 4L\big(\mathbb{E}[f(w^t)|\mathcal{F}_s]-f(w^*)\big)+\frac{4L}{n}\sum_{j=1}^n\sum_{\tau=s}^tp_{\tau}\big(\mathbb{E}[f_j(w_{\tau})|\mathcal{F}_s]-f_j(w^*)\big)\\
    &= 4L\big(\mathbb{E}[f(w^t)|\mathcal{F}_s]-f(w^*)\big)+4L\sum_{\tau=s}^tp_{\tau}\big(\mathbb{E}[f(w_{\tau})|\mathcal{F}_s]-f(w^*)\big),
\end{aligned}
\end{equation}
here $\{\mathcal{F}_t\}_{t\ge 0}$ is the filtered probability space, $t-T\le s\le t$ (recall $T$ is the length of inner iteration) is the latest available full gradient time stamp, $p_\tau$ is the probability distribution of stored gradient discussed in (10). Since $t-s$ is upper bounded (this is true for SVRG/SAGA++, as to SAGA, the expectation is $n\log n$ by ``\textit{Coupon collection problem}''), together with linear convergence, we know the second term is close to the first term up to a constant.
\subsection{Proof of Theorem 6}
First of all, we have the following recursive formula:
\begin{equation}
\begin{aligned}
P_g(x, \eta, c, n)&=\mathsf{Prox}_g(P_g(x, \eta, c, n-1)-c)\\
&=
\begin{cases}
P(x, \eta, c, n-1)-c-\eta, & \text{if } P(x,\eta, c, n-1)\ge c+\eta\\
0, &\text{if } c-\eta\le P(x, \eta, c, n-1) \le c+\eta\\
P(x, \eta, c, n-1)-c+\eta, & \text{if } P(x, \eta, c, n-1) \le c-\eta
\end{cases}.
\end{aligned}
\end{equation}
Because $c$ can be either positive or negative but $\eta$ is always positive, we consider about following cases:
\begin{itemize}
    \item ($c<-\eta$) In this case $0>c+\eta>c-\eta$, if:
    \begin{enumerate}
        \item $x\ge c+\eta$, then $P(x, \eta, c, n)=x-n(c+\eta)$;
        \item $x< c+\eta$, then suppose $x=q(c-\eta)+\epsilon$, $q\in\mathbb{N}$, $\epsilon\in [c-\eta, c+\eta]$, if $q\ge n$ then $P(x,\eta,c,n)=x-n(c-\eta)$; else $P(x,\eta, c, q)=\epsilon$, $P(x,\eta, c, q+1)=0$, $P(x, \eta, c, n)=-(n-q-1)(c+\eta)$.
    \end{enumerate}
    \item ($c>\eta$) In this case $0<c-\eta<c+\eta$ which is symmetric to previous case, if:
    \begin{enumerate}
        \item $x\le c-\eta$, then $P(x,\eta, c, n)=x-n(c-\eta)$;
        \item $x>c-\eta$, then suppose $x=q(c+\eta)+\epsilon$, $q\in\mathbb{N}$, $\epsilon\in[c-\eta, c+\eta]$, if $q\ge n$ then $P(x,\eta, c, n) = x-n(c-\eta)$; else $P(x,\eta, c,q)=\epsilon$, $P(x,\eta, c, q+1)=0$, $P(x,\eta, c, n)=-(n-q-1)(c-\eta)$.
    \end{enumerate}
    \item ($-\eta\le c\le \eta$) finally, $c-\eta\le 0\le c+\eta$, if:
    \begin{enumerate}
        \item $x\ge n(c+\eta)$, then $P(x,\eta, c, n)=x-n(c+\eta)$;
        \item $x\le n(c-\eta)$, then $P(x,\eta, c, n)=x+n(c-\eta)$;
        \item otherwise, $\lfloor \frac{x}{c+\eta} \rfloor<n$ or $\lfloor \frac{-x}{-c+\eta} \rfloor<n$ then we know it will eventually be zero: $P(x, \eta, c, n)=0$.
    \end{enumerate}
\end{itemize}
Clearly this is a piecewise linear function with tangent either $1$ or $0$.

\subsection{\label{app:l2reg}$\ell_2$ Logistic Regression Experiment}
\begin{figure}[htb]
    \centering
    \includegraphics[width=0.9\linewidth]{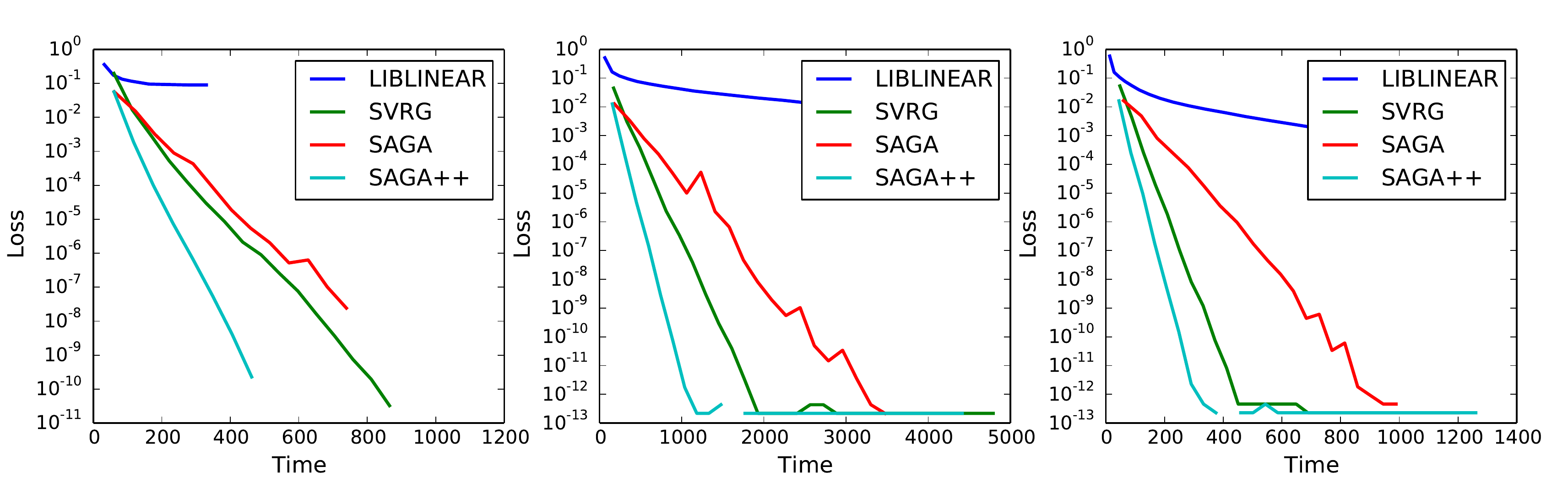}
    \caption{\label{fig:l2reg}Running time comparison among different data ($\lambda=1.0\times 10^{-7}$ for all data).}
\end{figure}

In this supplemental experiment, we conduct the $\ell_2$ logistic regression experiment, formulated as follows
\begin{equation}
    \label{eq:l2reg}
    w^*=\mathop{\arg\min}_w \frac{1}{n}\sum_{i=1}^n\log\big(1+\exp(y_ix_i^\intercal w)\big) + \frac{\lambda}{2}\Vert w \Vert^2_2.
\end{equation}
The datasets and settings are the same as $\ell_1$ experiment discussed in the main text. The experiment result is exhibited in Figure~\ref{fig:l2reg}.

\end{document}